\documentclass[runningheads]{llncs}
\usepackage{graphicx}
\usepackage{amsmath,amssymb} 

\usepackage{threeparttable}
\usepackage{booktabs}
\usepackage{multirow}
\begin{document}

\title{Better Guider Predicts Future Better:\\ Difference Guided Generative Adversarial Networks} 
\titlerunning{DGGAN Predicts Future } 


\author{Guohao Ying\inst{1} \and
Yingtian Zou\inst{2} \and
Lin	Wan \thanks{Lin	Wan is the corresponding author (e-mail: wanlin@hust.edu.cn).} \inst{1} \and
Yiming	Hu\inst{1} \and
Jiashi	Feng\inst{2}
}
%

\authorrunning{Guohao Ying, Yingtian Zou et al.} 


\institute{Huazhong University of Science and Technology, China \and
National University of Singapore
}

\maketitle

\begin{abstract}
Predicting the future is a fantasy but practicality work. It is the key component to intelligent agents, such as self-driving vehicles, medical monitoring devices and robotics. In this work, we consider generating unseen future frames from previous observations, which is notoriously hard due to the uncertainty in frame dynamics. While recent works based on generative adversarial networks (GANs) made remarkable progress, there is still an obstacle for making accurate and realistic predictions. In this paper, we propose a novel GAN based on inter-frame difference to circumvent the difficulties. 
More specifically, our model is a multi-stage generative network, which is named the Difference Guided Generative Adversarial Network (DGGAN). The DGGAN learns to explicitly enforce future-frame predictions that is guided by synthetic inter-frame difference. Given a sequence of frames, DGGAN first uses dual paths to generate meta information. One path, called Coarse Frame Generator,  predicts the coarse details about future frames, and the other path, called Difference Guide Generator, generates the difference image which include complementary fine details. Then our coarse details will then be refined via guidance of difference image under the support of GANs. With this model and novel architecture, we achieve state-of-the-art performance for future video prediction on UCF-101, KITTI.
\end{abstract}
\section{Introduction}
Predicting the future has drawn increasing attention due to its great practical value in various artificial intelligence applications, such as guiding unmanned vehicles, monitoring patient condition, to name a few. In this paper, we consider the  task that learns from the prior video frames to predict the future frames. Some previous distinguished works which aim to predict the low-level information like action~\cite{rezazadeganreal}, the flow~\cite{jin2017predicting,sedaghat2016next}, or skeleton~\cite{villegas2017learning,barsoum2017hp} have shown remarkable success. The limitation is that they do not predict the holistic information. To ameliorate it, we adopt this strategy which develop a  model that can acquire complete future RGB-images of the future not just one-sided information. The images can then be transferred to other video analysis tasks such as action recognition or utilized for models based on reinforcement learning.

However, the generation of realistic frames is a challenging task, especially when it is required to generate the whole foreground/background and unambiguous motion dynamics. Intuitively, for the sake of obtaining accurate prediction under the precondition of realistic generation, one has to delve deeper into previous adjacent frames. Some existing methods directly generate future frames by encoding context information using generators like CNNs~\cite{mathieu2015deep}, LSTM~\cite{byeon2017fully,lotter2016deep} Auto-Encoder~\cite{patraucean2015spatio} or GANs~\cite{liang2017dual,bhattacharjee2017temporal,mathieu2015deep,vondrick2016generating}. Unfortunately, those methods often suffer from blurry problem. To alleviate the issue, a more elegant method is 
to introduce a motion field layer under the assistance of auxiliary information. Those layer can produce motion dynamics, which transform pixels from previous frames to future frames~\cite{liu2017video}.
practical end-to-end fashions~\cite{villegas2017learning,liang2017dual} usually incorporate auxiliary information
such as optical-flow, skeleton-information through neural networks. They acquire complete clear frames from  combination of the previous frames and auxiliary information. Nevertheless, complicated loss to control the matrix transformation is ineluctable when most of above models aim at transform the given frames to future frames. Following this inspiration, guiding by an efficient auxiliary information is the keystone and a easier implement method is better.
Consequently, we aim at developing a better ``guider'' that predicts future more accurately and relieve the blurry problem.

To acquire this better guider, we resort to a strong motion information map, the Inter-Frame Difference Image. In particular, we propose the Difference Guided Generative Network(DGGAN) model that learns to generate the predicted frames and the difference frames which encode the difference between adjacent frames.  To get over the hurdle of blurry problem, we deploy the predicted inter-frame difference  as the guider for future frame prediction. Combining the guider with the previous frames, in that way we can apply pixel-shift from prior distributions instead of generating images from random-noise. the proposed model finally obtains the future frames with fine details and proper smoothness. Our end-to-end trainable model deploying this strategy achieves the state-of-the-art performance on multiple video benchmark datasets without complicated computing on transformation.

In summary, our main contributions are three-folds:
\begin{figure}[t]
\centering
\includegraphics[width=12cm]{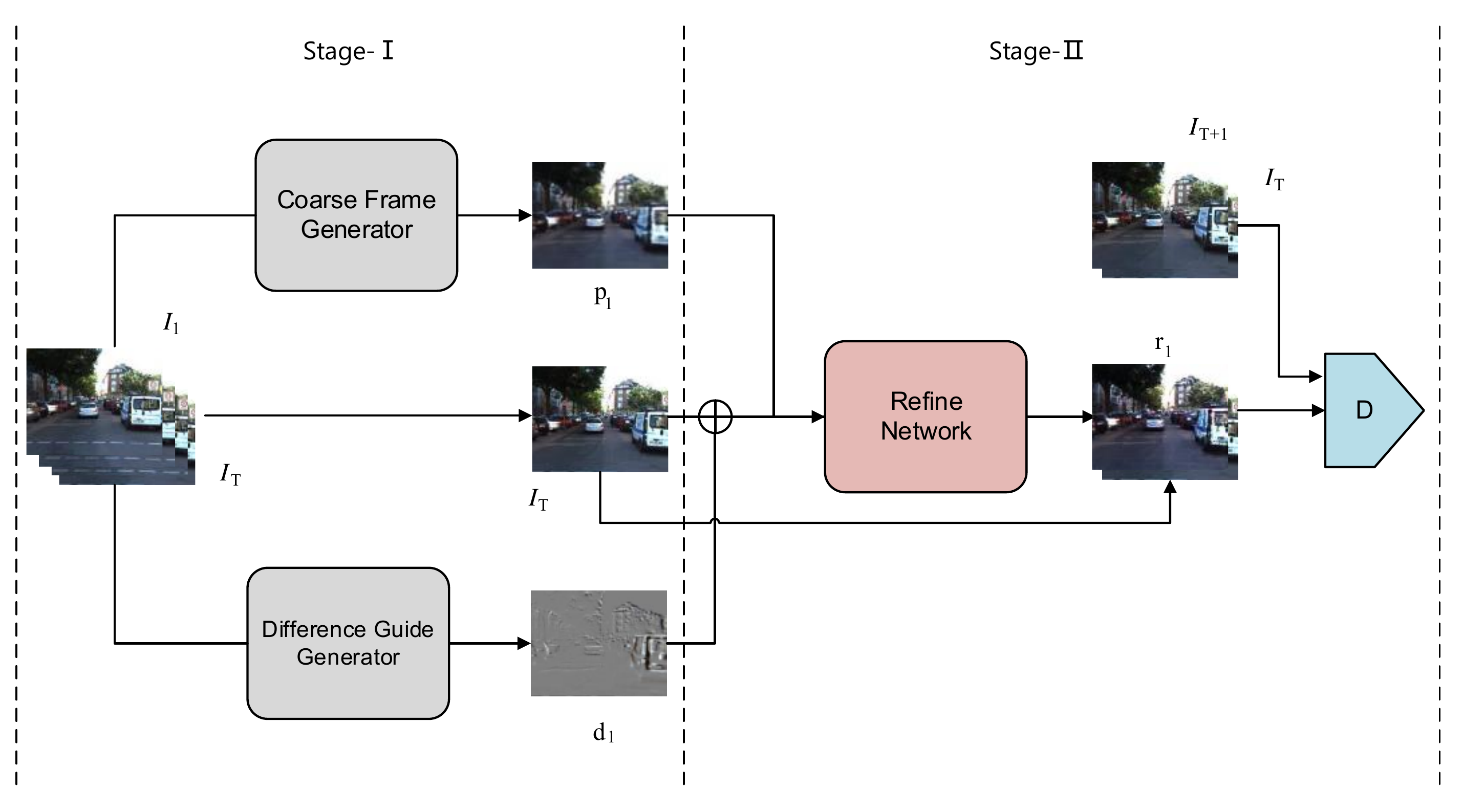}
\caption{Overview of Difference Guide Generative Adversarial Network (DGGAN).  The Coarse~Frame~Generator generates the coarse result of $T+1$ frame, $p_1$. Difference Guide Generator generates the difference image $d_1$ of $I_{T+1}$ and $I_{T}$. During stage-\uppercase\expandafter{\romannumeral2}, $p_1$ will be concatenated with $I_{T}$ "$\oplus$" by $d_1$ and then input to Refine Network where "$\oplus$" means pixel-wise sum. A discriminator will adversarial train the Refine Network to output the final result $r_1$ which is the fine prediction of $T+1$ frame.}
\label{fig_architecture}
\end{figure}
\begin{itemize}
\item We proposed DGGAN model which can generate inter-frame difference image and complete future frame. In single future image prediction, DGGAN achieve the state-of-the-art on UCF-101, KITTI. 

\item Inter-frame difference is a neglected powerful guider in motion video analysis. To our best knowledge, we are the first to introduce it into future prediction where the guider compels motion information become dominant. 

\item  Learning from~\cite{denton2015deep,bhattacharjee2017temporal} we propose a multi-stage model which will lessen the mistiness after single stage. And we experimentally demonstrated that our Refine~Network is efficacious on improving the coarse prediction generated by GAN. 

\end{itemize}


\section{Related works}
Video prediction has been more popular since it can be used in various of areas including self-driving, medical monitoring, or robotics. One way to make a simple prediction is predicting the future frames.
Subsequently we will introduce some prominent existing Network Architectures for future frames prediction.  

Video frame prediction is a challenging task due to the complex appearance and motion dynamics of natural scenes. Early approaches only use the RGB-frames~\cite{byeon2017fully,lotter2016deep} as input information, using CNN, RNN or LSTM to handle multi-frames input, and generate at least one future frame.  Further notice, as Generative Adversarial Networks~\cite{goodfellow2014generative} showing good performance compared with traditional generating works,~\cite{mathieu2015deep,liang2017dual,vondrick2016generating,ohnishi2017hierarchical} utilize GANs to achieve better generation. To obtain more effective information from previous scenario, they also take low-level features such as optical flows or separating the foreground and background as auxiliary information.

A series of works above attempt to propose more effective networks to  generate the future frames pixel directly. But the output is often blurry and training is costly, which usually spends a long time. An alternative approach to alleviate those problems is copying pixels from previous frames~\cite{liang2017dual,patraucean2015spatio,jin2017predicting,villegas2017learning}. Patraucean $et~al.$~\cite{patraucean2015spatio} use optical flows to encode a grid to transform the front one frame to the next one frame~\cite{patraucean2015spatio}. And the work of Jin $et~al.$~\cite{jin2017predicting} define a Voxel Flow which is coded from the previous frames to transform the front frames to the new one. Liang $et~al.$~\cite{liang2017dual} use a dual Motion GAN  to generate optical flow and frames, they make use of the optical flow to warp the last one frame from the input frame-sequences to the next new one frame. At the same time, they use the generated frame connected with the last one frame to  encode optical flow. They have two discriminators, which are used to control the generating of optical flow and frames~\cite{liang2017dual}. Xiong et al.~\cite{xiong2018learning} apply multiple GANs on raw frames for generation  and refinement. Although most of the aforementioned networks have good performance when predicting future frames, they may suffer from sophisticated transformation and need complicated loss for controlling the images generation. Another contemporary model of Liang $villegas.$ make use of the fusion of skeleton and previous adjacent frames instead of sophisticated transformation and need complicated loss. But the limitation of this model is that it can be only utilized for the prediction of foreground including different human actions, and the variation of background is hard to predict. Zhao et al.~\cite{zhao2018learning} use highly specific motion information such as 3DMM based face expression motion and human body keypoints motion for  face and human  video generation. In this sense, their method~\cite{zhao2018learning} is cumbersome than as it needs to manually select  motion cues   for generating videos of different themes. 

Our networks makes full use of the advantages of the above model and  avoids their weakness as much as possible. Our approach exploit simple loss to make the machine learn the coarse predicted frame and Difference Guide(DG) information which contain the variation of foreground and background from the previous frames to the new frames. Then the fusion of DG and previous adjacent frames can help getting over the problem of blurry. In addition our ultimate prediction is refined via GANs. To achieve more stable and superior training results, we use WGAN-GP~\cite{gulrajani2017improved} in combination with CGAN ~\cite{mirza2014conditional} instead of original GANs.

\section{Difference Guided GAN}
As mentioned above, predicting clear and accurate future frames has been attracting lots of attention. Though several works begin to employ various credible auxiliary information as the guide~\cite{patraucean2015spatio,ohnishi2017hierarchical,villegas2017learning,liang2017dual}, their performance is still not satisfactory. In this work, as shown in Figure~\ref{fig_architecture}, we devise a multi-stage generative network based on a better guide\textemdash the difference image,  to effectively overcome their limitations.
Concretely, during stage-\uppercase\expandafter{\romannumeral1}, we introduce the dual-path networks\textemdash one path contains the Coarse Frame Generator and  the other one contains the Difference Guide Generator. The upper path generates the coarse result of future predicted frame while the lower path generates the difference image between last frame and predicted frame which we regard it as the guide. To adequately learn the context, we set multi-frames as the input sequence. where the recent frames will serve as constraints. Besides, we use additional losses to constrain the generation of difference image and coarse predicted frame. At stage-\uppercase\expandafter{\romannumeral2}, the coarse predicted frame, difference image and last frame will be fused by Refine Network. There is a discriminator that adversarially trains the Refine Network to enhance reality of the synthetic image. Reasons for this design are: (1) difference image compels motion information become dominant; (2) under the guidance of difference motion information, we could refine our obscure results. In this way, we could get clear and accurate predicted image.     
\begin{figure}[t]
\centering
\includegraphics[width=12cm]{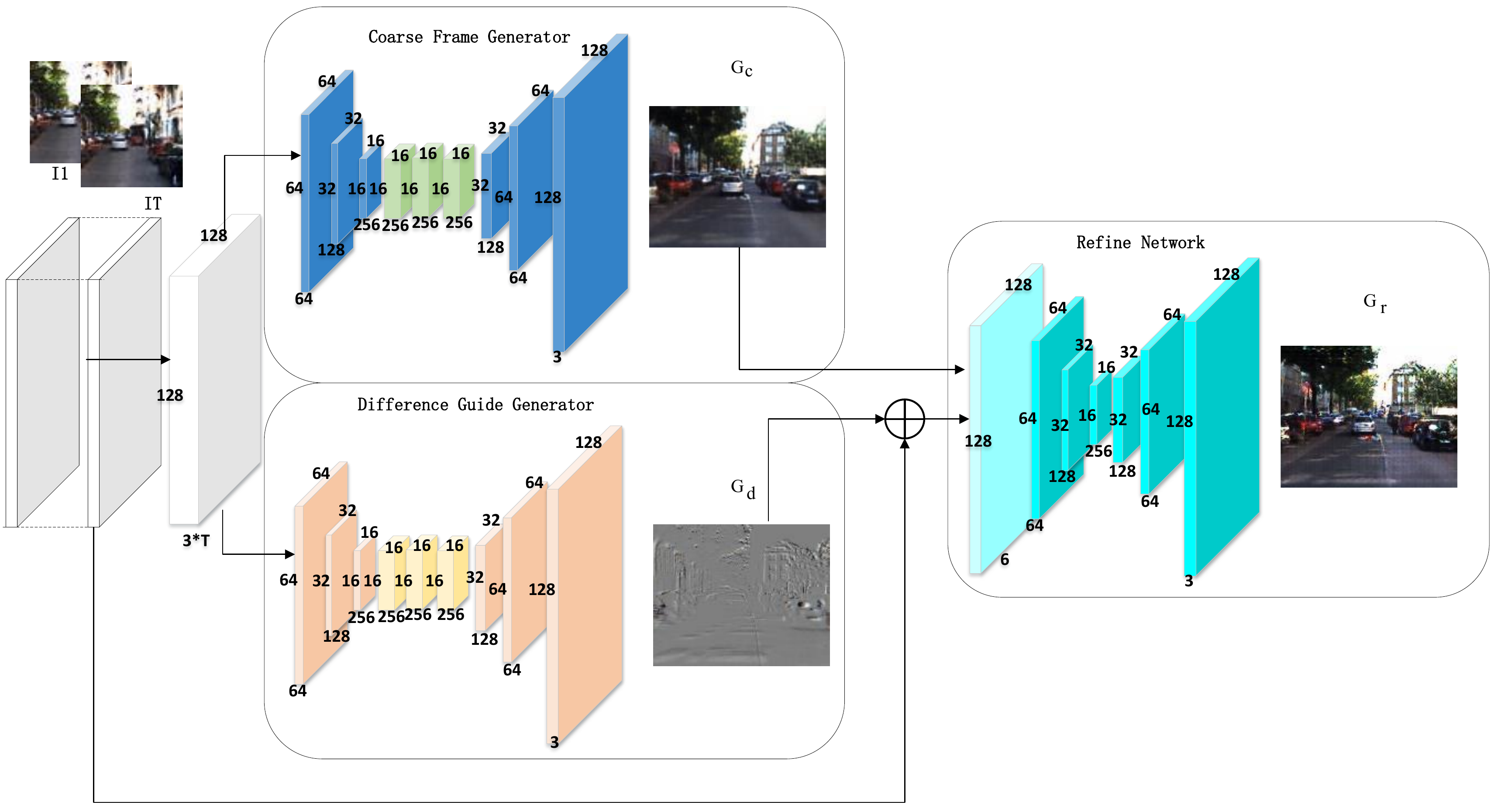}
\caption{Frameworks of generators. Firstly, the sampled frames sequence of length $T$ are contacted together, where each frame contain 3 channels. There are two input paths that identical sequence is fed into Coarse Frame Generator(CFG) and  Difference Guide Generator(DGG) simultaneously. The CFG and DGG have the same structure which includes three convolutional layers, three res-block layer, and three deconvolutional layers. But the different generation target that produce Coarse Frame $G_{c}$ and Difference Image $G_{d}$ respectively. Then pixel-wise adding $G_{d}$ and $I_{T}$ to get $\tilde{G}_d$. $\tilde{G}_d$, $G_{c}$ will be fed into Refine Network(RN). Finally, the RN refine those afferent features to ultimate accurate prediction $G_{r}$. }
\label{fig_generators}
\end{figure}
\begin{figure}[h]
\centering
\includegraphics[width=8cm]{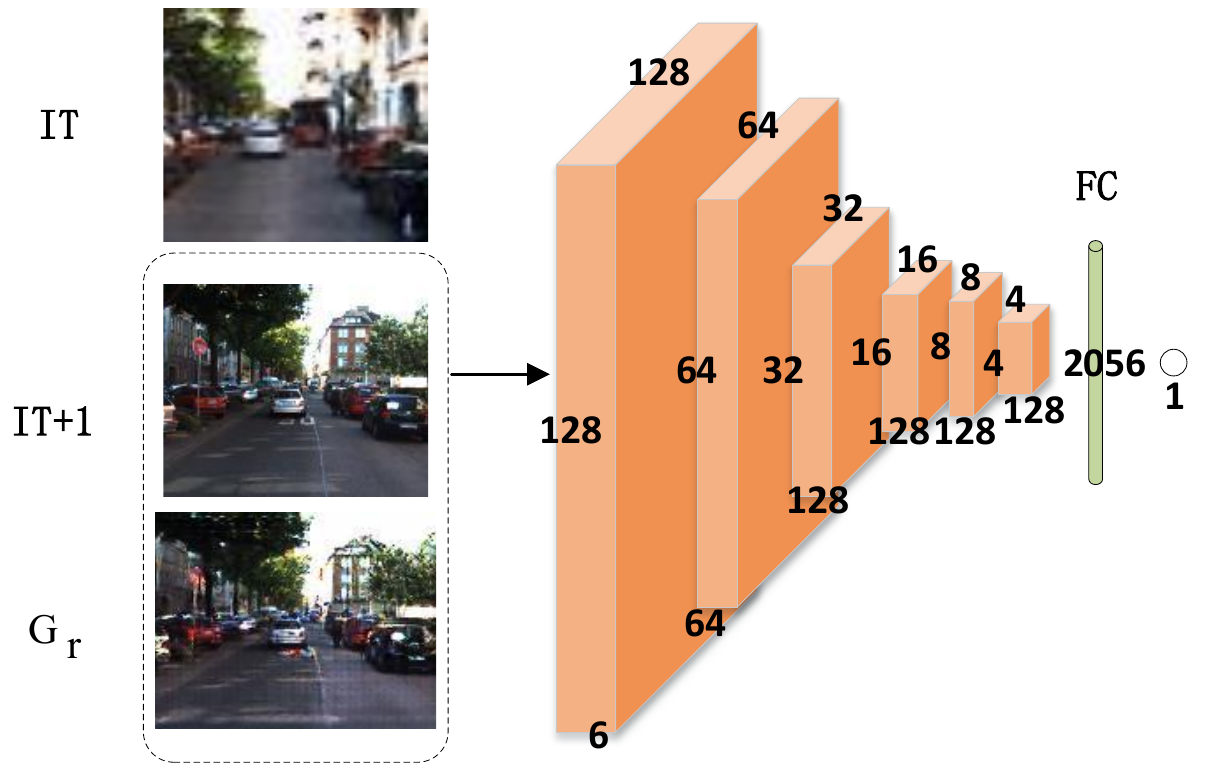}
\caption{Frameworks of discriminator. Taking $I_{T+1}$ or $G_{r}$ contacted with $I_{T}$} as input. 
\label{fig_generators}
\end{figure}

\subsection{Coarse Frame Generator}
As shown in Fig.~\ref{fig_architecture}, the  Coarse Frame Generator(CFG) in the upper path predicts coarse information in future frames. 
CFG is structurally similar to the Transform Net in~\cite{johnson2016perceptual}, which contains an encoder, a res-network and a decoder. The encoder contains three convolution layers, each of which is  followed by a Batch-norm layer~\cite{ioffe2015batch} and a LeakyReLU~\cite{maas2013rectifier} layer. In contrary to encoder, the decoder has three deconvolution layers, and all the convolution or deconvolution layers utilize 4$\times$4 filters and stride 2 to process input feature maps. At the end of decoder, we add $tanh$ activation function for normalization.
More formally, let $S_I=\{I_1,I_2,\ldots,I_T\}$ denotes the input  sequence  of  $T$ frames and $I_{T+1}$ denotes the future frame.  After encoder, we can get features $O_1 = f_{conv}(S_I)$. With only three convolution layers, $O_1$ is a poor representation in such a difficult task. As each $S_I$ contains consecutive frames which may include the same $I_t$, such as $S_I=\{I_1,I_2,\ldots,I_T\}$ and $\hat{S}_I=\{I_2,I_3,\ldots,I_{T+1}\}$. Besides, each $I_t$ have similar features. Thus if we deepen the net, inconspicuous variation in consecutive frames may cause the gradient vanishing in training phaze. Under these circumstances, we insert the res-networks~\cite{he2016deep} between the encoder and decoder. The res-networks include three residual blocks and each block contains three convolution layers which use simple 1$\times$1 filters and stride 1 to process input feature maps., followed by a Batch-norm layer~\cite{ioffe2015batch} and a ReLU. When passing through the blocks, we can get feature maps $O_2 = f_{res}(O_1)$ where $f_{res}(O_1) = \mathcal{F}(O_1, \{Wi\}) + O_1$. Deconvolution layers in decoder will then up-sample $O_2$ to be the same size with input frames. In a nutshell, the CFG learns to produce coarse predicted frame $G_{c}$.  

At this stage, we simplify the awkward task that we don't need to predicted the ultimate clear and realistic frame in one step. At the beginning, DGGAN only need to learn to fit the distribution of raw data. The fitting is important for providing structural information for our last refined prediction. There, we use the Mean Square Error(MSE) loss function to train the CFG: 
\begin{equation}
\mathcal{L}_{cf}(G_{c},I_{T+1})  = \sum\limits_{i\in I} \left( G_{c}(i) -I_{T+1}(i)\right)^2,
\end{equation}
where $G_{c}(i)$ and $I_{T+1}(i)$ means the $i$-th pixel value of coarse predicted frame $G_{c}$ and real predicted frame $I_{T+1}$. 

\subsection{Difference Guide Generator}

In the lower path, the inter-frame difference is generated by Difference Guide Generator(DGG). As it shown in Fig.~\ref{fig_generators}, the identical input $S_I=\{I_1,I_2,...,I_T\}$ have two paths, lower of which will then be processed by the encoder and decoder of DGG. As for the encoder and decoder, DGG and CFG share the same structure. What distinguishes them is the different learning objective that target of DGG is generating Difference Image. First of all, we need to get real difference image $\hat{D}$, where $\hat{D}_{T} = I_{T+1}-I_{T}$. Through the DGG, we obtain the predicted difference image $G_{d}$. Generating this sparse image $G_{d}$ is an easier task compared to generating realistic image which is dense. Thus, we introduce the $L_1$ loss to constrain the generation which outperform the MSE Loss when dealing with the generation of sparse image:
\begin{equation}
\mathcal{L}_{dg}(G_{d},\hat{D}_{T})  = \sum\limits_{i\in~\hat{D}} \left |G_{d}(i) - \frac{\hat{D}_{T}(i)}{2}\right|
\end{equation}
where $G_{d}(i)$ and $\hat{D}_{T}(i)$ denote the $i$-th pixel value of generated difference image sequences $G_{d}$ and real difference image sequence $\hat{D}_{T}$. Note that we take $tanh$ activation function to normalize our output, but the ground-truth of difference image $\hat{D}_{T}$ is between $[-2,2]$. So we divide $\hat{D}_{T}$ by factor 2 to nondimensionalize the $\hat{D}_{T}$.
Difference image plays a guide role in the following stage. Contrary to generating predicted frames directly, generating the sparse motion information is easy to converge. This well-trained information will guide coarse prediction to produce more accurate result. 

For stage-I, the CFG and DGG generate coarse predictions $G_{c}$ and difference images $G_{d}$ respectively at each time step. A MSE penalty item constrains the CFG and $L_1$ penalty item constrains the DGG. In summary, the holistic penalty of stage-I is:
\begin{equation}
\label{eq_stage1}
\mathcal{L}_{stage-I} = \mathcal{L}_{cf}(G_{c},I_{T+1}) + \mathcal{L}_{dg}(G_{d},\hat{D} _{T})
\end{equation}
From the (\ref{eq_stage1}), it suggests that CFG and DGG are synchronously trained under their own objective. The targets of stage-I are making coarse result approach to the future frame and generating more accurate difference image.

\subsection{Refine Network}
After stage-\uppercase\expandafter{\romannumeral1}, we can obtain coarse predicted frame $G_{c}$ and difference guide $G_{d}$. As ``$\oplus$" shown in Figure~\ref{fig_generators}, the guide $G_{d}$ pixel-wise plus $I_{T}$ will synthesize the guided image $\hat{I}$ where $\hat{I}_{T+1} = G_{d} \times 2 + I_{T}$.

Coarse predicted frame $G_{c}$ have blurry problems which is far short of generating the realistic images. On the other side, $\hat{I}$ of combining the difference image and previous adjacent frame directly have abundant artifacts. Thus we define a Refine Network(RN) in stage-\uppercase\expandafter{\romannumeral2} to smooth the predicted frame. It has the analogic function with Auto-Encoder which learns to compress data from the input layer into a compact code, and then recover the original data from the code. To fuse the frame $\hat{I}_T$ and $G_{d}$, we set the guide as a condition. They are taken as the input of RN. RN merely has a simple structure which consists of three convolutional layers and three deconvolutional layers.
 
We add Batch-Norm layer and a LeayReLU layer following the convolution and deconvolution layer. And at the last output, we use $tanh$ activation function to normalize. After RN, the coarse prediction $G_{c}$ will be refined to a high visual quality prediction $G_{r}$ with reduced blurriness or artifacts.

\subsection{Adversarial Training}
Since the GANs~\cite{goodfellow2014generative} has achieved tremendous success on generating of realistic images, we adopt GAN-based training strategy to refine our coarse results. As the original GAN suffers from several training difficulties such as mode collapse and instable convergence. In DGGAN, we adopt WAGN-GP~\cite{gulrajani2017improved} as our learning tactic. The  WAGN-GP theoretically solve problems of original GAN mentioned before by minimizing an approximated Wasserstein distance and remedy the limitation of weight clipping about WGAN via the utilization of more advanced gradient penalty. In order to attain the goal of predicting, we also control our generation with the restrictions of given information like the principle of CGAN. As the Difference Image to be a condition guiding the generating, that's why we call Difference guide. As shown in Figure~\ref{fig_architecture}, discriminator(D Net) located at the end of model is used for training the refined results. D Net has five convolutional layers with 4$\times$4 filters and stride 2 to process input feature maps and one fully connection layers. Follow the WGAN-GP~\cite{gulrajani2017improved}, we use layer-norm layer instead of batch-norm layer. 

\textbf{Training of the discriminator.}
when traing the discriminator, we regard the refined output $G_{r}$ stacked with last frame $I_{T}$ as negative example where the $I_{T}$ is the condition. Analogously, we can get positive example by concatenate real image $I_{T+1}$ with $I_{T}$. The optimization objective of discriminator can be written as:
\begin{equation}
\begin{split}
\mathcal{L}_{adv}^{D} =&E_{r\sim\mathbb{P}(Gr|I_{T})} [D(r)] - 
E_{I\sim\mathbb{P}(I_{T+1}|I_{T})} [D(I)]\\
&+\lambda E_{\hat{x}\sim\mathbb{P}_{\hat{x}}} [(\lVert\nabla_{\hat{x}} D(\hat{x})\rVert_2-1)^2]
\end{split}
\end{equation}

where the $\mathbb{P}(I_{T+1}|I_{T})$ is the distributions of real future frames in combination with conditions, the $\mathbb{P}(Gr|I_{T})$ is the distributions of synthesized prediction in combination with conditions, $\hat{x}$ is the random samples between $I_{T+1}$ and $G_{r}$ and $\lambda$ is the coefficient. $\hat{x}$ helps us to circumvent tractability issues by enforcing such a soft version of the constraint with a penalty on the gradient norm which lessen the distraction and burden from generating realistic images. 

In this way, our discriminator learn to distinguish the ground-truth and refined generation with the condition of previous adjacent frame.

\textbf{Training of the RN.}
Contrary to discriminator, we regard the refined output $G_{r}$ stacked with last frame $I_{T}$ as positive input. keeping the weights of D fixed, and we perform an optimization on RN:
\begin{equation}
\mathcal{L}_{adv}^{RN} =-E_{r\sim\mathbb{P}(Gr|I_{T})} [D(r)]
\end{equation}

By minimizing the above two loss criteria (4), (5), RN trying to confuse the generated frames and real future frames. Meanwhile the the ability of D that distinguish the synthesized frames and the real future frames is promoting. At last, the D can no longer make sure the source of input frames, and the output prediction of RN will be realistic to the real future frames.

In summary, the optimization target of stage-\uppercase\expandafter{\romannumeral2} is:
\begin{equation}
\label{eq_stage1}
\mathcal{L}_{stage-\uppercase\expandafter{\romannumeral2}} = \mathcal{L}_{adv}^{RN} + \mathcal{L}_{adv}^{D}
\end{equation}

\subsection{Multi-frame Generation}
Our model can be competent at Multi-frame Generation. We take the generated frame $G_{r1}$ combined with previous $\{I_2,...,I_{T+1}\}$ as the new input sequence, then use it to generated $r_2$ which is closed to real-frame $I_{T+2}$. By repeating the above operation, we can acquire a sequence of frames $S_r = \{G_{r1},G_{r2},...,G_{rn}\}$ where $n$ is the length of multi-frames. 
Because all the following prediction are influenced by the first frame prediction, our main target is making the one-frame prediction as accurate as possible.
\section{Experiments}

\subsection{Experimental Set-up}
\textbf{Datasets.} 
We evaluate our proposed model in three different real-world scenarios, UCF-101 dataset~\cite{soomro2012ucf101}, KITTI dataset~\cite{geiger2013vision} and Human 3.6M dataset ~\cite{ionescu2014human3}. The UCF-101 dataset, which contains 13320 annotated videos, include many human's activities. It was split to three subsets. We take the first subset as our training set and testing set. In KITTI dataset, it includes many driving-scenarios from different road conditions. Training set and testing set are from two categories: road and cars. Human 3.6M dataset is formed by various videos which consists of plentiful motions of humans. We extract these video to frames as well. In those dataset, to produce examples, we extracted every 5 continuous frames each step. We give the front four frames as input, and then try to predict the next one frames which is similar to the last one. The data patches are firstly normalized to the range [-1:1] so that their values are equal to the generation interval of $tanh$.

\textbf{Quality evaluation.} To quantify the comparison with State-of-the-Art methods, we use Structural Similarity Index(SSIM)~\cite{wang2004image},  Mean Squared Error(MSE)~\cite{lotter2016deep} and Peak Signal-to-Noise Ratio(PSNR)~\cite{huynh2008scope} evaluation prototypes to assess the image quality of the results. SSIM is used for measuring the similarity between two images and higher value of SSIM  means better accuracy of predicting. The MSE represents the quality of predicted frames. It is always non-negative and value closer to zero is better. PSNR is similar to MSE, which can assess approximation to human perception of reconstruction quality, and higher value means better results. 

\textbf{Training details}
We use the Adam~\cite{kingma2014adam} to optimize DGGAN at the batch-size of 8 in both two stages. In stages-I, Both CFG and DGG use the same learning rate which is set to 0.001 and gradually decreased to 0.0001 over time. It takes about 100epochs to reach convergence.

In stage-II, keeping the weight of that networks in stages-I fixed, the learning rate about Refine Network and the discriminator is set to 0.0001 and gradually decreased to 0.00001 over time. It takes about 200 epochs to reach convergence. We also set the weighting parameter $\lambda$ = 10 in $L_{adv}^{D}$.

\subsection{Comparison with the State-of-the-Art}
As shown in Table~\ref{table_human}-\ref{table_kitti}, we list our results and other State-of-the-Art approaches in detail.

\textbf{Comparison on UCF-101.} Firstly, We assess our model on UCF-101. We equally spaced sample from videos and choose examples from part-one as train and test datasets. The frames are resized to 128$\times$128.  Table~\ref{table_ucf} displays the results compared with other methods.

We take the one-frame prediction experiment results illustrated in Deep Multi-stage(DMS) ~\cite{bhattacharjee2017temporal}. For multi prediction setting, there is no data reported in~\cite{liu2017video}. 

Note that in single-frame prediction, our network surpasses the state-of-the-art comprehensively and transcends Deep Voxel Flow(DVF) 2\% on SSIM and 10.6 on PSNR. For second frame prediction, DGGAN transcends the state-of-the-art 2\%, 5.6 on SSIM, PSNR respectively. To our knowledge, we have achieved the state-of-the-art on UCF-101 on these two different settings. In another point, our approach is similar to DVF who generates a motion information(Voxel Flow) to guide the input frames. But DGGAN guiding by Difference shows the better performance than DVF. That's we think Difference has a stronger lead than Vodel Flow.

\begin{table}[h]
\begin{center}
\begin{tabular}{|l|c|c|c|c|}
\hline
Method & GAN & SSIM(2nd frame) & PSNR(2nd frame)\\
\hline\hline
BeyondMSE~\cite{mathieu2015deep}& $\surd$ & 0.92(0.89) & 32(28.9) \\
DVF~\cite{liu2017video}& $\times$  & 0.96(-) & 35.8(-) \\
*DMS~\cite{bhattacharjee2017temporal}&$\surd$   & 0.95(0.93) & 38.2(36.8)\\ 
\hline

Ours & $\surd$ & \textbf{0.98}(\textbf{0.95})  &\textbf{46.4}(\textbf{42.3})\\
\hline
\end{tabular}
\end{center}
\caption{Comparison of performance for different methods using SSIM/PSNR scores for the UCF-101. Value in ``()" means score of second frame $r_{2}$. "*" This score is provided by~\cite{bhattacharjee2017temporal} which predicted four frames once a time, but we only compare with the first frame of it.  "$\surd$" means the model based on GAN.}
\label{table_ucf}
\end{table} 

\begin{figure}[h]
\centering
\includegraphics[width=12cm]{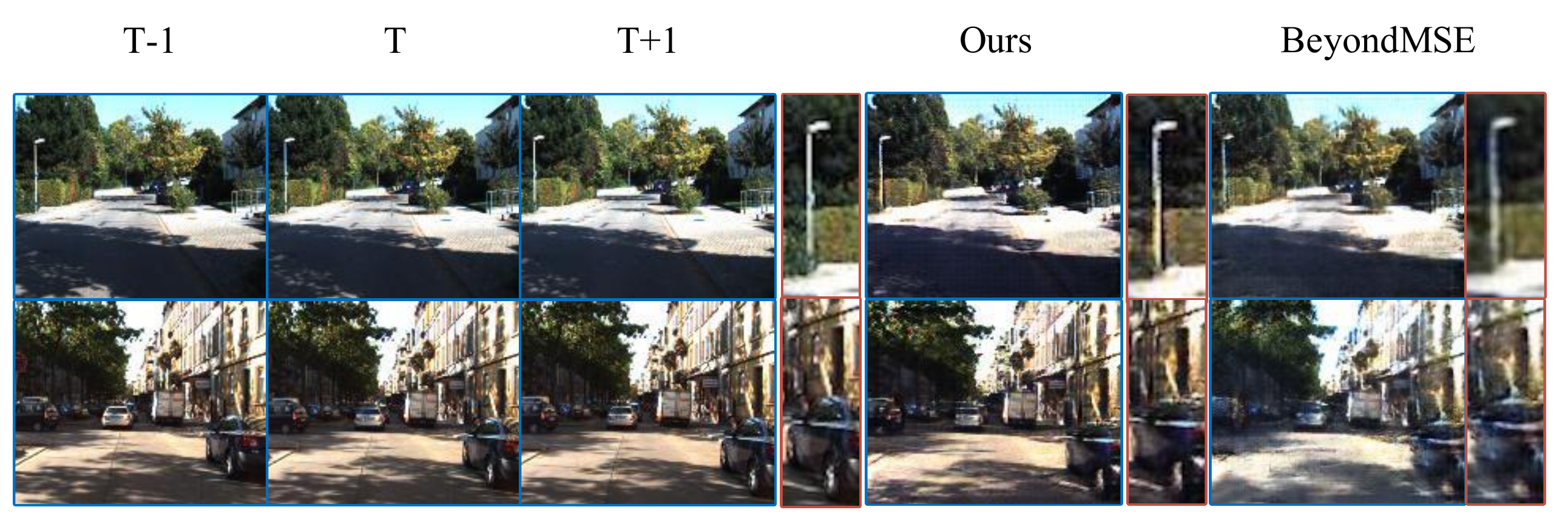}
\caption{Results in KITTI. From column 1-3 are past frames $T-1,~T$ and ground truth of frame $T+1$. Column 4 in blue is our prediction of frame $T+1$. Column 5 in blue is the result of BeyondMSE~\cite{mathieu2015deep}. These three narrow columns in red are zoom in details of the $T+1$ real frame, our prediction and result of BeyondMSE respectively.}
\label{fig_results}
\end{figure}

\textbf{Comparison on KITTI.} 
Secondly, we assess our model by training on the KITTI dataset and testing on the CalTech Pedestrian dataset from "city", "Residential", and "Road". Frames from both datasets are center-cropped and down-sampled to 128$\times$160 pixels. Different from previous dataset, the KITTI dataset have obvious background pixels changing. Nonetheless, our model can be capable of the prediction task. Table~\ref{table_kitti} shows all the consequence compared with other methods. All the score are provided by the Dual Motion Gan(DMG) ~\cite{liang2017dual}. Our model excess BeyondMSE~\cite{mathieu2015deep}, PredNet~\cite{liu2017video} and  DMG~\cite{liang2017dual} on SSIM. On MSE, we also have the minimum error. In particularly, compared with DMG which take optical flow as "guider", the results confirm the superiority of difference image. More intuitive performance are shown in Figure \ref{fig_results}. In spite that KITTI has greater variation between frames, we still anticipate the precise position about streetlight, cars and windows. And in comparison with BeyondMSE, our generated images are clearer significantly, and boost the image quality conspicuously.

\begin{table}[h]
\begin{center}
\begin{tabular}{|l|c|c|c|}
\hline
Method & SSIM & MSE$(\times10^{-3})$\\
\hline\hline
BeyondMSE~\cite{mathieu2015deep}  & 0.881 & 3.26\\
PredNet ~\cite{liu2017video}  & 0.884  & 3.13\\
DMG~\cite{liang2017dual}   & 0.899 & 2.41\\ 
\hline

Ours  & \textbf{0.902} & \textbf{2.18}\\
\hline
\end{tabular}
\end{center}
\caption{Comparison of  different methods using SSIM/MSE scores for the KITTI.}
\label{table_kitti}
\end{table} 

\textbf{Comparison on Human 3.6M.} We assess our model in Human 3.6M at last. We extract frames from video and choose 100000 frames which contain a mass of consecutive motions randomly as the training set, and we take 10\% of entire dataset as testing set. All of frames are down-sampled to 64$\times$64. As seen in Table~\ref{table_human}, it reports the quantitative comparison with the state-of-the-art methods from BeyondMSE~\cite{mathieu2015deep}, DNA~\cite{finn2016unsupervised} and Full Context(FC)~\cite{byeon2017fully}. We re-implement the result of BeyondMSE according to \cite{mathieu2015deep} that minimizes the loss
functions in BeyondMSE(ADV+GDL) under the same setting with our model. And the result about DNA and FC refer to the the experimental results shown in their papers. Since the dynamic regions in Human 3.6M are the central human movings, and usually the background is static, the difference guide have shown improvement in local variation of pixels. Under the guiding of it, our model significantly outperform BeyondMSE. And because DNA and FC need ten frame as input but our model just use four frame as input, we can also compete within them in a comprehensive way.  

\begin{table}[h]
\begin{center}
\begin{tabular}{|l|c|c|c|}
\hline
Method & SSIM & PSNR\\
\hline\hline
BeyondMSE~\cite{mathieu2015deep} & 0.90 & 26.7\\
DNA~\cite{finn2016unsupervised} & 0.992 & 42.1\\
FC~\cite{byeon2017fully} & \textbf{0.996} & \textbf{45.2}\\
\hline

Ours  & 0.990 & 44.1\\
\hline
\end{tabular}
\end{center}
\caption{Comparison of different methods using PSNR/SSIM scores for the Human 3.6M.}
\label{table_human}
\end{table}

\begin{figure}[t]
\centering
\includegraphics[width=8cm]{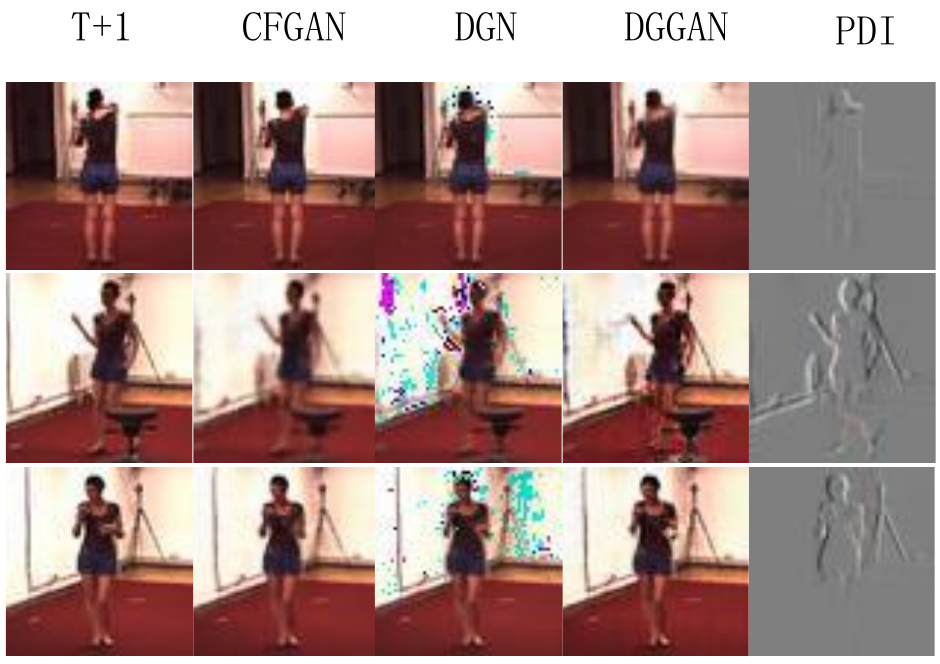}
\caption{Visual presentation of one-frame prediction on Human 3.6M. From left to right are ground truth of frame T+1, results of Coarse Frame Generative Adversarial Network(CFGAN), Difference Guided Net(DGN), our proposed model(DGGAN) and Predicted Difference Image(PDI).}

\label{human}
\end{figure}

\begin{figure}[h]
\centering
\includegraphics[width=12cm]{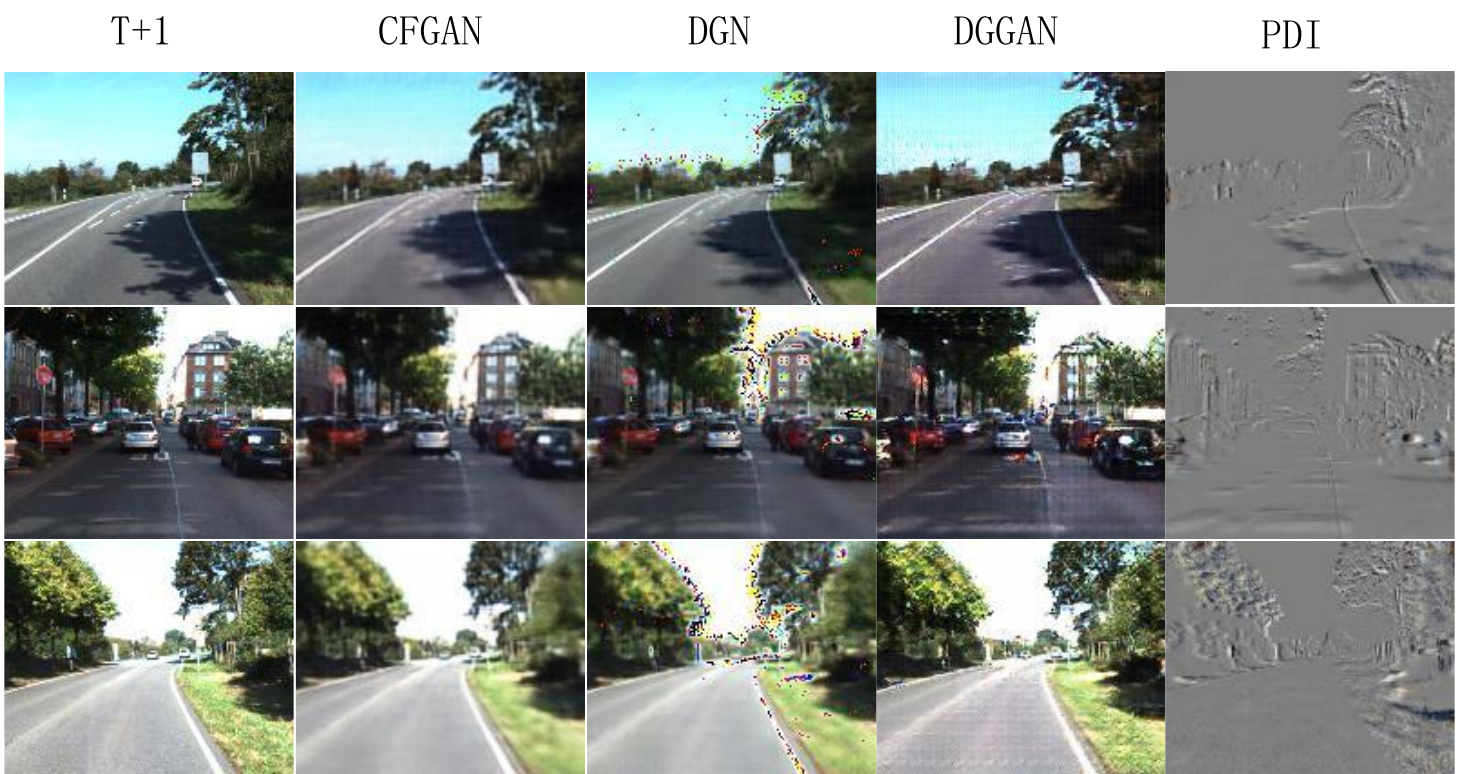}
\caption{Visual presentation of one-frame prediction on KITTI. The label means the same thing as Fig.\ref{human}}

\label{kitti}
\end{figure}

\begin{figure}[h]
\centering
\includegraphics[width=12cm]{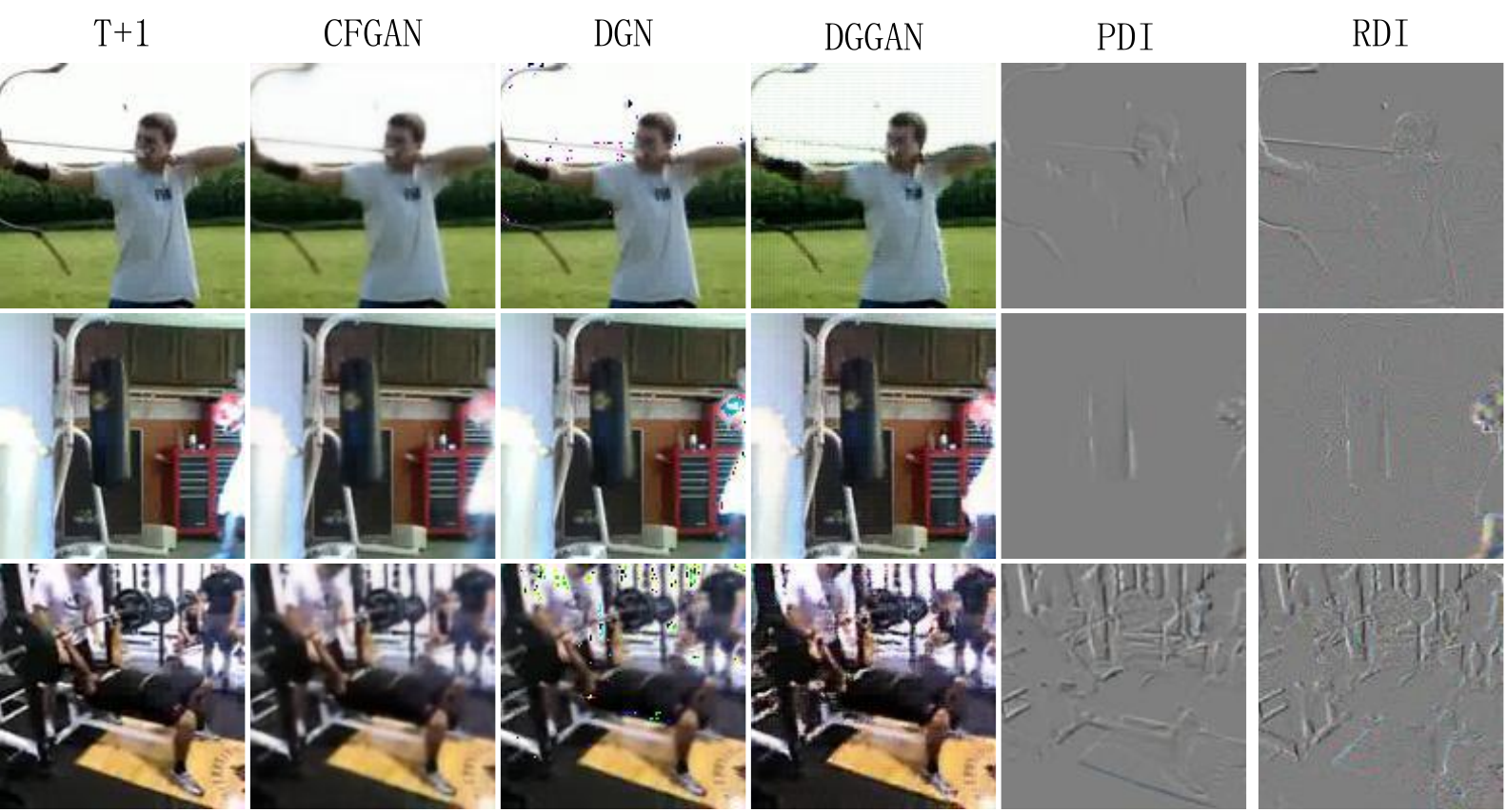}
\caption{Visual presentation of one-frame prediction on UCF-101. The label means the same thing as Fig.\ref{human}. Besides, Real Difference Image(PDI) means the ground truth of difference image between $I_T$ and $I_{T+1}$}

\label{ucf-101}
\end{figure}

\subsection{Evaluation on effectiveness}

\begin{table}[t]

  \begin{threeparttable}
  \caption{Comparisons of our ultimate network with other two path methods in stage-I on Human 3.6M, UCF-101 and KITTI. The order of magnitude in "MSE" column is $(\times10^{-3})$ .}  
  \label{tab:performance_cmc}

    \begin{tabular*}{\textwidth}{@{\extracolsep{\fill}}ccccccccccc}
    \toprule  

    \multicolumn{2}{c}{\multirow{2}{*}{Model}}
    &\multicolumn{3}{c}{Human 3.6M}&\multicolumn{3}{c}{UCF-101}&\multicolumn{3}{c}{KITTI}\cr

   	\cmidrule(lr){3-5}\cmidrule(lr){6-8}\cmidrule(lr){9-11}
    &&SSIM&PSNR&MSE&SSIM&PSNR&MSE&SSIM&PSNR&MSE\cr  
    \midrule 

\multicolumn{2}{c}{Copy}
    &0.90&-&-&0.80&-&-&0.67&-&-\cr
    \multicolumn{2}{c}{CFGAN}
    &0.87&29.4&3.9&0.89&31.5&4.2&0.79&28.4&5.26\cr
  \multicolumn{2}{c}{DGN}
    &0.97&41.6&2.1&0.96&44.7&2.9&0.86&35.6&2.9\cr
    \multicolumn{2}{c}{DGGAN}
    &\textbf{0.99}&\textbf{44.1}&\textbf{1.8}&\textbf{0.98}&\textbf{46.4}&\textbf{2.4}&\textbf{0.90}&\textbf{37.5}&\textbf{2.1}\cr

    \bottomrule  
    \end{tabular*} 
 
    \label{table_evaluation}
    \end{threeparttable}  
\end{table}

In order to evaluate our model, we set three baselines to prove the effectiveness of DGGAN in Table~\ref{table_evaluation}. In general, the copy of last frame $I_T$(Copy) is a significant reference. Without the guide, we append a discriminator to upper path which only contains the CFG as the second baseline named Coarse Frame Generative Adversarial Network(CFGAN). For third baseline, we directly apply the guide $G_{d}$ into the last frame $I_{T}$ to generate the prediction. There is a striking enhancement when comparing the Copy(Row 1) and DGN(Row 3). Accordingly, the difference image's intrinsic motion guidance is ideally suited for future prediction.

Drill down further to analysis the Table~\ref{table_evaluation}, the gap between DGGAN and CFGAN tells that multi-stage and dual path GAN is more effective than plain GAN. Especially on KITTI, conspicuous variation between inter-frames,  copying pixels from last frame(Row 1) have a poor performance than DGN(Row 3) that indicates it is vital to utilize motion guider. Furthermore, DGGAN also exceeds three benchmarks on all the evaluation indexes. The results of these contrast experiments illustrate that our proposed DGGAN has a reasonable structure. Both on guiding the motion information prediction and refining the synthetic image, DGGAN has shown its potential and superiority in this field.

More Intuitive performance are shown in Fig \ref{human}-\ref{ucf-101}. Owing to the hard task that CFGAN need to generate the whole future frames directly, it's  inevitable that generated frames are a bit blurry. on the contrary, our DGGAN focus on the variations between inter-frames under the guide of difference images and show better performance on generation. Beyond that, the similarity of Real Difference Image(PDI) and Predicted Difference Image(PDI) show the validity of task that using L1 loss to control the generation of sparse difference image. In addition, as we can see, the use of difference images directly will produce the images with lots of artifact. and the RN can solve this problem by refining the images effectively.

\section{Conclusion}
In future prediction, we proposed a novel and reasonable methodology, Difference Guide Generative Adversarial Network(DGGAN). This method can refine the synthetic predicted image under the guiding of difference image. Although recent works provide plenty of alternative strategies such as leveraging optical flow to guide the prediction. DGGAN still stands out through a better guide. To explore effectiveness of DGGAN, we conducted a serials of experiments on comparing the state-of-the-art and our benchmarks. As we expected, DGGAN experimentally and theoretically demonstrated its excellent capacity which could be further applied to action analysis and video generation.

%
%
%
\newpage
\bibliographystyle{splncs04}
\bibliography{egbib}

\end{document}